\documentclass[letterpaper]{article}
\usepackage[preprint]{aaai2027}

\usepackage[hyphens]{url}
\usepackage{graphicx}
\usepackage{natbib}
\usepackage{caption}
\usepackage{amsmath}
\usepackage{amssymb}
\usepackage{booktabs}
\usepackage{algorithm}
\usepackage{algorithmic}
\usepackage{newfloat}
\usepackage{listings}

\DeclareCaptionStyle{ruled}{labelfont=normalfont,labelsep=colon,strut=off}
\floatstyle{ruled}
\newfloat{listing}{tb}{lst}{}
\floatname{listing}{Listing}

\title{Correlation-Aware Evidential Fusion for Trusted Multi-View Classification}
\author{
    Tianyang Wang,
    Mengchi Zuo,
    Cai Xu\corresponding,
    Yilin Zhang,
    Ziyu Guan,
    Wei Zhao
}
\affiliations{
    School of Computer Science and Technology, Xidian University, China\\
    \{tianyangwang@stu., mczuo@stu., cxu@, ylzhang\_3@stu., zyguan@, ywzhao@mail.\}xidian.edu.cn
}

\begin{document}

\maketitle

\begin{abstract}
Trusted multi-view classification aims to make accurate and reliable predictions by integrating information from multiple views. Existing methods typically estimate class evidence and uncertainty for each view and combine the resulting opinions through aggregation rules. However, they generally treat different views as independent information sources, overlooking the correlations among them. Highly correlated views tend to produce similar evidence, and treating them as independent sources may repeatedly count consistent information and give it disproportionate influence, while weakening complementary evidence that provides additional information. Consequently, such correlation-unaware fusion may not only limit classification performance but also produce miscalibrated uncertainty estimates. To address this issue, we propose Correlation-Aware Fusion of Evidence (CAFE), a framework for trusted multi-view classification. CAFE applies complementarity-based trust discounting to the evidence produced by view-private experts and introduces a correlation-aware constraint to reduce the repeated contribution of similar evidence. It then learns sample-specific weights from the global multi-view context to adaptively combine the correlation-adjusted private evidence with cross-view collaborative evidence. Extensive experiments on six real-world datasets demonstrate that CAFE consistently improves classification and calibration performance in general settings, with particularly significant gains when strong correlations exist among views.

\end{abstract}


\section{Introduction}

Multi-view classification aims to improve classification performance by integrating information from multiple views
\cite{wei2026unbiased,he2026bootstrapping}.
It has been widely applied in key areas such as autonomous driving \cite{wei2023surroundocc},
medical diagnosis
\cite{chen2022pathomic,chen2022pancancer},
and multimedia content understanding
\cite{clark2026molmo2}.
However, in safety-critical applications, accurate predictions alone are insufficient: models should also quantify the reliability of their outputs \cite{gawlikowski2023survey}.

Trusted multi-view classification addresses this issue by estimating both class support and predictive uncertainty \cite{han2021trusted}. Recent methods are largely built on evidential deep learning and subjective logic. Typically, each view-specific network produces non-negative class evidence, which parameterizes a Dirichlet distribution. This is then converted into an opinion consisting of class-wise belief masses and uncertainty. These opinions are aggregated by averaging, weighting, or Dempster--Shafer combination to obtain the final prediction and uncertainty. Building on this, existing methods address practical challenges including view conflict
\cite{xu2024reliable},
noise
\cite{xu2024labelnoise},
missing views
\cite{xie2023exploring},
and variations in view quality
\cite{liu2025enhancing}.

\begin{figure}[t]
\centering
\includegraphics[width=\columnwidth]{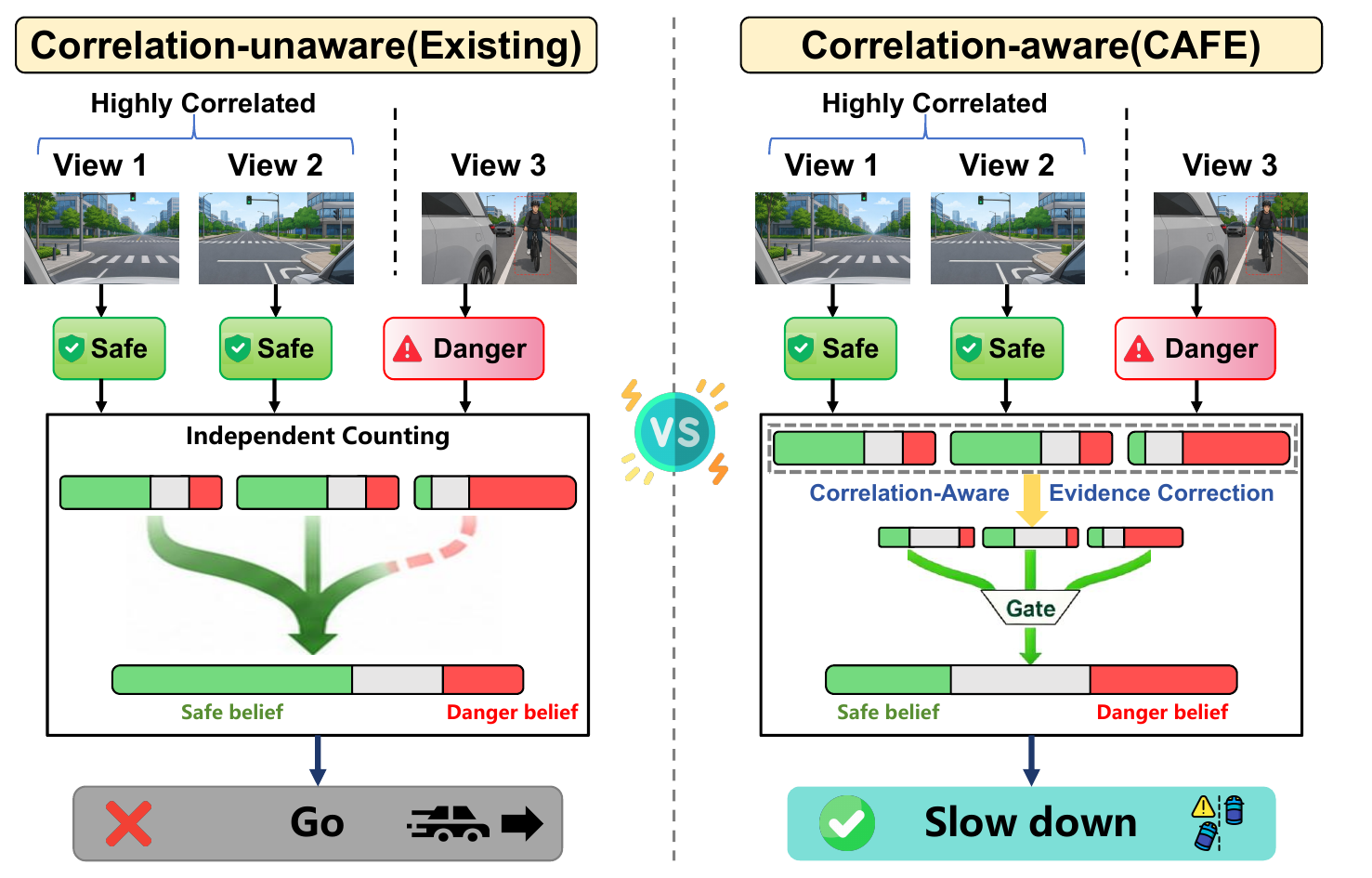}
\caption{Impact of highly correlated views on decision fusion in existing methods and CAFE. CAFE appropriately discounts redundant information while preserving complementary information.}
\label{fig:motivation}
\end{figure}

Despite using different aggregation strategies, existing methods follow a correlation-unaware evidential fusion paradigm. They typically treat evidence from different views as independent and ignore inter-view correlations, which may lead to negative effects. For example, as shown in Figure~\ref{fig:motivation}, views~1 and~2 capture similar forward-facing scenes and produce consistent evidence for \emph{Safe}, whereas view~3 detects a cyclist from a different perspective and provides complementary evidence for \emph{Danger}. This causes two negative effects: (1) performance degradation. correlation-unaware fusion may repeatedly reinforce the correlated evidence for \emph{Safe}, overwhelm the risk evidence from the view~3, and produce an incorrect decision; (2) uncertainty miscalibration. Accumulating correlated evidence can distort fused evidence strength: confident correlated views may lead to overconfidence, whereas uncertain ones may suppress complementary support and lead to underconfidence.

These observations motivate our exploration of the correlation-aware evidential fusion paradigm. We formulate three core requirements for this paradigm: explicitly representing view-specific and cross-view evidence, correcting redundant contributions from correlated views, and learning adaptive fusion weights across training samples rather than relying on predefined rules. To this end, we propose CAFE, a correlation-aware evidential fusion method for trusted multi-View classification. CAFE first constructs view-private experts to preserve view-specific evidence and a cross-view collaborative expert to capture complementary information across views. It then employs complementarity-based discounting and correlation-aware regularization to suppress redundant private evidence and correct its effective strength. Finally, a global-context gate jointly evaluates the corrected private and collaborative evidence and assigns sample-specific fusion weights, yielding more accurate decisions and better-calibrated uncertainty estimates.

Our main contributions are summarized as follows:
\begin{itemize}
    \item We revisit the problem of view correlation in multi-view learning from an evidence-level perspective. We show that treating evidence from correlated views as independent information leads to redundant evidence accumulation, which further affects both classification decisions and uncertainty calibration.
    \item We propose CAFE, which integrates view-private and cross-view collaborative experts. Complementarity-based trust discounting and correlation-aware regularization suppress repeated evidence contributions and correct evidence strength.
    \item Experiments under standard and strongly correlated settings show that CAFE achieves the highest accuracy on all six datasets, with 2.4\% higher average accuracy and 11.9\% lower average ECE than the strongest baselines. Under strongest correlation, its accuracy drops by only 3.4\%, versus 7.5\%--35.7\% for representative baselines.
\end{itemize}




\section{Related Work}

\subsection{Evidential Multi-View Learning}

Reliable decision-making requires predictive confidence to reflect actual correctness, especially in safety-critical applications.Evidential deep learning provides an alternative deterministic approach to uncertainty estimation. EDL learns class-specific evidence through a single forward pass and uses the resulting evidence to parameterize a Dirichlet distribution, thereby representing both class beliefs and overall uncertainty\cite{sensoy2018evidential}. Building on this foundation, TMC extends evidential deep learning to multi-view classification and employs the Dempster–Shafer rule to combine view-specific opinions\cite{han2021trusted}, while ECML further develops an opinion aggregation strategy for conflicting views\cite{xu2024reliable}. Recent studies have paid increasing attention to differences among view-specific evidence in terms of contribution, confidence estimation, and strength scale. These methods perform adaptive fusion by assessing the evidence contribution of each view\cite{xu2025beyond}, improve confidence estimation using a distribution-aware uncertainty measure\cite{liang2025trusted}, or alleviate the incomparability of evidence scales across views through unified expert routing\cite{zhang2026independently}. However, they largely overlook dependencies among evidence from correlated views. Our method models such decision-level dependencies to reduce repeated accumulation and calibrate fused evidence strength.

\subsection{Correlation-Aware Multi-View Learning}
Modeling view correlations is an important research direction in multi-view learning. Early studies were primarily based on canonical correlation analysis, which laid the foundation for learning shared representations by identifying projection spaces that maximize the correlation between different views\cite{hotelling1992relations}. Building on this foundation, one line of research combines correlation analysis with deep neural networks, autoencoders, or tensor decomposition to capture nonlinear, reconstruction-preserving, or high-order cross-view correlations\cite{andrew2013deep,wang2015deep,benton2019deep}.

Recent studies have shifted from learning shared representations to jointly modeling cross-view consistency and complementarity. One line of methods preserves shared information across views and view-specific information through intact-space learning, shared-specific representation decomposition, or disentangled feature learning
\cite{xu2015multiview,xu2021multivae,chen2025dfl}. Another line explicitly constructs view relationships based on cross-view topology, anchor-graph structures, feature correlations, or sample structures, and uses these relationships to guide representation enhancement and adaptive fusion
\cite{dong2023crossview,liang2025view,liu2024learn,li2026miscc}. Although these methods improve multi-view representations and fusion, they mainly model correlations at the feature level, leaving dependencies among decision evidence underexplored. In contrast, we model decision-level correlations to suppress repeated contributions and calibrate evidence strength.

\begin{figure*}[t]
\centering
\includegraphics[width=\textwidth]{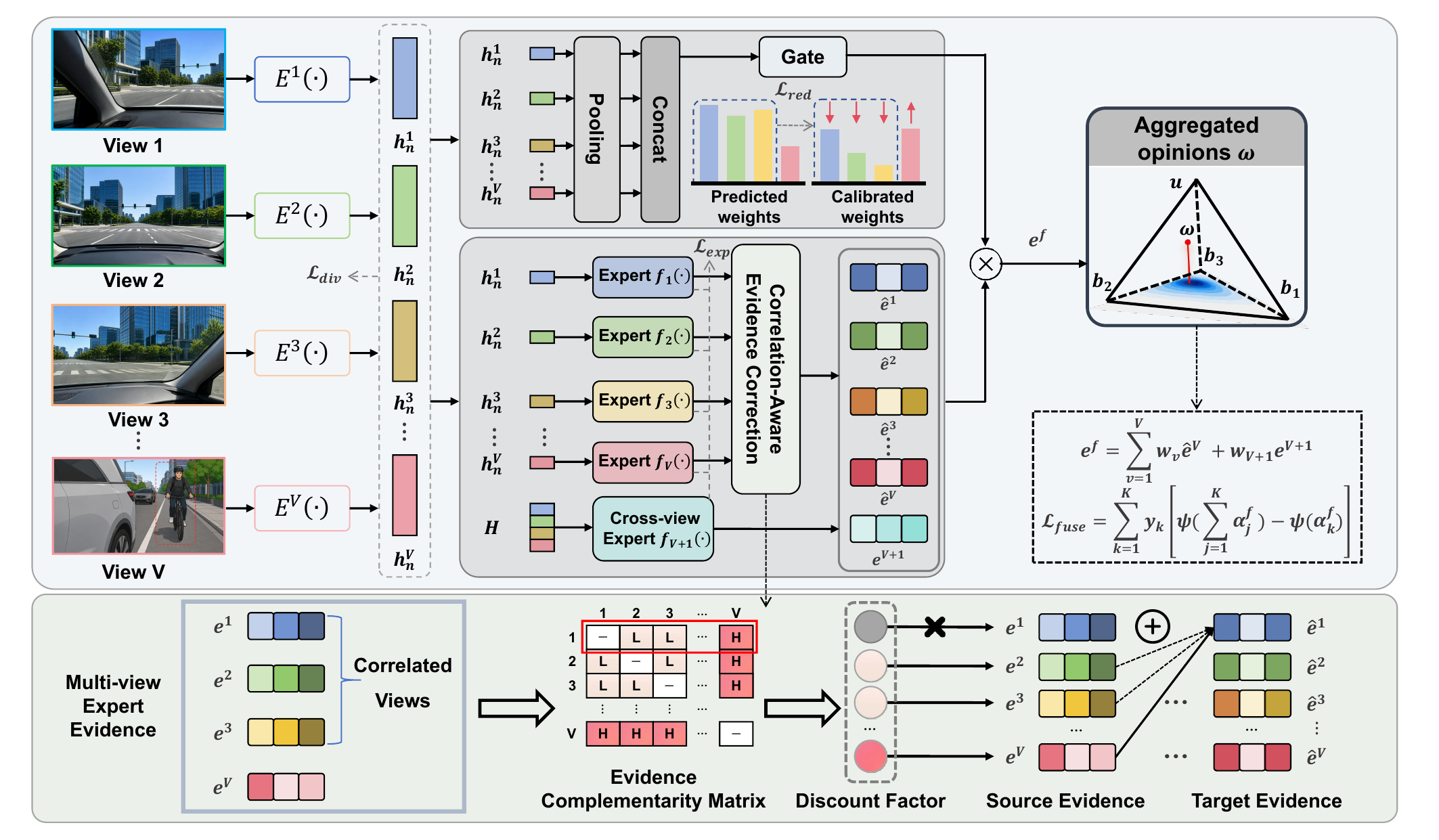}
\caption{Overall framework of CAFE. Each view is first mapped into a unified representation space. The view-private experts and the cross-view collaborative expert extract view-specific and joint evidence, respectively. The correlation-aware evidence correction module discounts highly similar private evidence according to pairwise evidence complementarity. Finally, sample-specific expert weights derived from the global multi-view context are used to aggregate the corrected private evidence and collaborative evidence, producing the final prediction and uncertainty estimate.}
\label{fig:cafe_framework}
\end{figure*}

\section{The Method}

\subsection{Problem Definition}

For a $K$-class multi-view classification problem, consider a dataset
$\mathcal{D}=\{\{\mathbf{x}_{n}^{v}\}_{v=1}^{V},\mathbf{y}_{n}\}_{n=1}^{N}$
with $N$ instances and $V$ views, where $\mathbf{x}_{n}^{v}\in
\mathbb{R}^{d_v}$ denotes the feature vector of the $n$-th instance
from the $v$-th view, and $d_v$ is the corresponding feature
dimension. The one-hot vector $\mathbf{y}_{n}\in\{0,1\}^{K}$ denotes
the ground-truth label of the $n$-th instance.

Our goal is to integrate information from all views to accurately predict
the class label $\mathbf{y}_{n}$ while estimating the associated predictive
uncertainty $u_n\in[0,1]$. A larger $u_n$ indicates greater uncertainty
in the prediction.

\subsection{The CAFE Pipeline}

Ignoring view correlations during evidence fusion may cause redundant evidence to be repeatedly accumulated, thereby reducing classification accuracy and producing miscalibrated uncertainty estimates. To address these issues, we propose CAFE, whose overall architecture is illustrated in Figure~\ref{fig:cafe_framework}. CAFE consists of three main components: mixed evidential experts, correlation-aware evidence correction, and evidence fusion with a global-context gate. The following subsections describe each component in detail.

\subsubsection{Evidential Expert Construction}

To preserve view-specific information while capturing shared information across views, we construct view-private experts and a cross-view collaborative expert. Each private expert retains the distinctive evidence provided by its corresponding view, while the collaborative expert captures joint cross-view evidence. Since raw features may have different dimensionalities, we employ view-specific projection layers to map them into a unified $d$-dimensional representation space:
\begin{equation}
\mathbf{h}^{v}
=
\operatorname{LN}
\left(
\mathbf{W}^{v}\mathbf{x}^{v}+\mathbf{b}^{v}
\right),
\qquad v=1,\ldots,V,
\label{eq:view_projection}
\end{equation}
where $\mathbf{W}^{v}$ and $\mathbf{b}^{v}$ are learnable parameters
for the $v$-th view, and $\operatorname{LN}(\cdot)$ denotes layer
normalization. 

For each view, we construct a view-private evidential expert
$f_v(\cdot)$ that takes the corresponding view representation as input
and produces class-wise evidence:
\begin{equation}
\mathbf{e}^{v}
=
\sigma_{+}
\left(
f_v(\mathbf{h}^{v})
\right),
\qquad v=1,\ldots,V,
\label{eq:private_evidence}
\end{equation}
where $\sigma_{+}(\cdot)$ denotes the activation function, which ensures non-negative evidence. These view-private experts preserve the decision information provided by individual views.

In addition to private-view information, combining multiple views may
reveal joint information that cannot be fully captured by any
individual view. We therefore concatenate all view representations
along the feature dimension to obtain a joint multi-view
representation:
\begin{equation}
\mathbf{H}
=
\left[
\mathbf{h}^{1};
\mathbf{h}^{2};
\ldots;
\mathbf{h}^{V}
\right].
\label{eq:joint_representation}
\end{equation}

The joint representation is then fed into a cross-view collaborative
expert $f_{V+1}(\cdot)$ to produce joint cross-view evidence:
\begin{equation}
\mathbf{e}^{V+1}
=
\sigma_{+}
\left(
f_{V+1}(\mathbf{H})
\right).
\label{eq:collaborative_evidence}
\end{equation}

Unlike the view-private experts that operate on individual views, the
collaborative expert models decision information arising from
interactions among all views.

Following evidential deep learning \cite{sensoy2018evidential}, the
evidence produced by any expert
$m\in\mathcal{M}=\{1,\ldots,V+1\}$ is used to parameterize a
Dirichlet distribution:
\begin{equation}
\boldsymbol{\alpha}^{m}
=
\mathbf{e}^{m}+\mathbf{1}.
\label{eq:expert_dirichlet}
\end{equation}

The Dirichlet strength, class-wise belief mass, and overall uncertainty
of the $m$-th expert are respectively defined as
\begin{equation}
S^{m}
=
\sum_{k=1}^{K}\alpha_k^{m},
\qquad
b_k^{m}
=
\frac{e_k^{m}}{S^{m}},
\qquad
u^{m}
=
\frac{K}{S^{m}},
\label{eq:expert_opinion}
\end{equation}
which satisfy
\begin{equation}
\sum_{k=1}^{K}b_k^{m}+u^{m}=1.
\label{eq:opinion_constraint}
\end{equation}

Accordingly, each expert produces a subjective opinion
$\boldsymbol{\omega}^{m}=(\mathbf{b}^{m},u^{m})$.
The resulting collection of expert evidence is denoted by
\begin{equation}
\mathcal{E}
=
\left\{
\mathbf{e}^{1},\ldots,\mathbf{e}^{V+1}
\right\},
\label{eq:expert_evidence_set}
\end{equation}
where $\mathbf{e}^{1},\ldots,\mathbf{e}^{V}$ are the view-private
evidence and $\mathbf{e}^{V+1}$ is the cross-view collaborative
evidence. 

\subsubsection{Correlation-Aware Evidence Correction}

The preceding section produces $V$ private-expert evidence vectors
$\{\mathbf{e}^{v}\}_{v=1}^{V}$ and the collaborative-expert evidence
$\mathbf{e}^{V+1}$. Correlated views may generate highly similar private
evidence, and directly fusing these vectors can repeatedly count redundant
decision information, overwhelm complementary evidence, and distort the fused
evidence strength. We therefore introduce a correlation-aware evidence
correction mechanism before final fusion to adjust the effective contribution
of each private expert according to pairwise evidence complementarity.

For any two private experts $i$ and $j$, their sample-level complementarity is 
defined as
\begin{equation}
c_{ij}
=
\sum_{k=1}^{K}
\left|e_k^i-e_k^j\right|.
\end{equation}

A larger $c_{ij}$ indicates that the two experts provide more distinct
information, whereas a smaller $c_{ij}$ indicates that they produce highly
similar evidence. Based on the relative complementarity, the discount factor
of expert $j$ with respect to expert $i$ is defined as
\begin{equation}
p_{ij}
=
\frac{2c_{ij}}
{\sum_{m\neq n}c_{mn}}.
\end{equation}

When two experts produce similar evidence, a smaller $p_{ij}$ imposes a
stronger discount on the opinion of expert $j$. Conversely, a larger $p_{ij}$
preserves more evidence that provides additional information.

Given the belief masses $\mathbf{b}^{j}$ and uncertainty $u^{j}$ obtained in 
the preceding section, the discounted opinion transmitted from expert $j$ to 
expert $i$ is expressed as
\begin{equation}
\widetilde{b}_k^{j\rightarrow i}=p_{ij}b_k^j,
\qquad
\widetilde{u}^{j\rightarrow i}
=1-\sum_{k=1}^{K}\widetilde{b}_k^{j\rightarrow i}.
\end{equation}

This operation proportionally reduces the belief masses and transfers the 
discounted portion to uncertainty, thereby decreasing the effective strength 
of correlated evidence.

After evidence discounting, each private expert $i$ acts as a local fusion center. Every other private expert $j\neq i$ transmits its discounted opinion $\widetilde{\boldsymbol{\omega}}^{j\rightarrow i}$ according to their complementarity. Each center then aggregates the transmitted opinions from the other $V-1$ private experts.

Given two transmitted opinions 
$\boldsymbol{\omega}^{A}=(\mathbf{b}^{A},u^{A})$ and 
$\boldsymbol{\omega}^{B}=(\mathbf{b}^{B},u^{B})$, we employ an 
uncertainty-weighted operator $\Diamond$ to fuse them:
\begin{equation}
\begin{aligned}
b_k^{A\Diamond B}
&=
\frac{
b_k^{A}(1-u^{A})u^{B}
+
b_k^{B}(1-u^{B})u^{A}
}{
u^{A}+u^{B}-2u^{A}u^{B}
}, \\[1mm]
u^{A\Diamond B}
&=
\frac{
(2-u^{A}-u^{B})u^{A}u^{B}
}{
u^{A}+u^{B}-2u^{A}u^{B}
}.
\end{aligned}
\end{equation}

This operator adjusts the local fusion contribution of each transmitted 
opinion according to its uncertainty, assigning greater influence to opinions 
with lower uncertainty.

For each local fusion center $i$, the discounted opinions transmitted from the 
other $V-1$ private experts are aggregated as
\begin{equation}
\widehat{\boldsymbol{\omega}}^{\,i}
=
\mathop{\Diamond}_{j\neq i}
\widetilde{\boldsymbol{\omega}}^{j\rightarrow i}
=
\left(
\widehat{\mathbf{b}}^{\,i},
\widehat{u}^{\,i}
\right).
\end{equation}

This process is performed for all private experts, yielding $V$ locally corrected opinions. Correlated opinions receive stronger discounting, reducing their belief masses and increasing uncertainty, thus suppressing their contributions to local aggregation. In contrast, more complementary opinions retain greater effective belief.

The corrected opinions are subsequently mapped back into the evidence space:
\begin{equation}
\widehat{e}_k^{\,i}
=
\frac{
K\widehat{b}_k^{\,i}
}{
\widehat{u}^{\,i}
}.
\end{equation}

\subsubsection{Global Context-Guided Evidence Fusion}

After evidence correction, the relative contributions of private and collaborative experts still vary across samples. Predefined fusion rules cannot learn such weighting patterns across training samples or adapt them to the current multi-view information. We therefore introduce a global-context gate that learns sample-specific expert weights from the global multi-view representation.

Let $\mathbf{h}^{v}$ denote the latent representation of the $v$-th view after 
projection. We first apply mean pooling to all view representations to obtain 
the global multi-view context of the current sample:
\begin{equation}
\bar{\mathbf{h}}
=
\frac{1}{V}
\sum_{v=1}^{V}\mathbf{h}^{v}.
\end{equation}

For the $v$-th private expert, its view representation $\mathbf{h}^{v}$ is 
concatenated with the global context $\bar{\mathbf{h}}$. The resulting 
representation is then fed into a scoring network $G_p(\cdot)$ shared by all 
private experts to calculate its score:
\begin{equation}
s_v
=
G_p\left(
[\mathbf{h}^{v};\bar{\mathbf{h}}]
\right),
\qquad
v=1,\ldots,V.
\end{equation}

Since the collaborative expert models joint cross-view information, its score 
is generated from the global context using an independent scoring network 
$G_c(\cdot)$:
\begin{equation}
s_{V+1}
=
G_c\left(
\bar{\mathbf{h}}
\right).
\end{equation}

This design allows each private-expert score to depend on its local view information and the global context across all views, while the collaborative-expert score is determined by the overall multi-view context. We collect the expert scores as $\mathbf{s}=[s_1,\ldots,s_{V+1}]$ and apply a temperature-scaled Softmax to obtain their weights:
\begin{equation}
\mathbf{w}
=
[w_1,\ldots,w_{V+1}]
=
\operatorname{Softmax}
\left(
\frac{\mathbf{s}}{\tau}
\right),
\end{equation}
where $\tau$ is a temperature parameter controlling the smoothness of the 
expert-weight distribution. After normalization, all expert weights are 
non-negative.

Finally, the corrected private-expert evidence and the uncorrected 
collaborative-expert evidence are fused using the sample-level weights:
\begin{equation}
\mathbf{e}^{f}
=
\sum_{v=1}^{V}
w_v\widehat{\mathbf{e}}^{\,v}
+
w_{V+1}\mathbf{e}^{V+1}.
\end{equation}

\subsection{Training Objective}

To ensure accurate fusion and informative, non-redundant expert evidence, the objective combines fused-evidence and average-expert supervision with private-expert diversity and correlation-aware weight regularization. The supervision terms optimize fused and expert predictions, while the regularizers preserve diversity and suppress repeated contributions from correlated experts.

For simplicity, let $\mathbf e^{(v)}$ denote the evidence of the $v$-th 
expert used in the final fusion. For $v=1,\ldots,V$, let
$\mathbf e^{(v)}=\widehat{\mathbf e}^{\,v}$ denotes the corrected evidence 
of the $v$-th private expert, while 
$\mathbf e^{(V+1)}=\mathbf e^{V+1}$ denotes the uncorrected evidence of the 
collaborative expert. The corresponding Dirichlet parameters are defined as 
$\boldsymbol{\alpha}^{(v)}=\mathbf e^{(v)}+\mathbf 1$.

\noindent\textbf{Fused evidence supervision.}
Since the final prediction is produced from the fused evidence, we employ 
the evidential Digamma loss to supervise the fusion output:
\begin{equation}
\mathcal L_{\mathrm{fuse}}
=
\sum_{k=1}^{K}
y_k
\left[
\psi\left(
\sum_{j=1}^{K}\alpha_j^f
\right)
-
\psi\left(
\alpha_k^f
\right)
\right],
\end{equation}
where $\boldsymbol{\alpha}^f=\mathbf e^f+\mathbf 1$ denotes the Dirichlet parameters of the fused evidence. Moreover, $\psi(\cdot)$ denotes the Digamma
function, and $\mathbf y$ is the one-hot label vector.

\noindent\textbf{Average expert supervision.}
Fused-output supervision alone may leave individual experts insufficiently discriminative, forcing the weighting network to compensate through fusion weights. We therefore apply the same evidential classification loss to all $V+1$ experts and average their losses:
\begin{equation}
\mathcal L_{\mathrm{exp}}
=
\frac{1}{V+1}
\sum_{v=1}^{V+1}
\sum_{k=1}^{K}
y_k
\left[
\psi\left(
\sum_{j=1}^{K}\alpha_j^{(v)}
\right)
-
\psi\left(
\alpha_k^{(v)}
\right)
\right].
\end{equation}

This supervision ensures that the corrected private and collaborative experts produce discriminative evidence for reliable adaptive fusion.



\noindent\textbf{Private-expert diversity regularization.}
Since the private experts are jointly optimized, their internal representations may gradually become similar. Let $\mathbf z^v=f_v(\mathbf h^v)$ denote the hidden representation produced by the $v$-th private expert before its evidence output layer. To preserve view-specific discriminative information, we introduce a diversity regularizer over these representations. Specifically, let $\widehat{\mathbf z}^{\,v}=\mathbf z^v/\|\mathbf z^v\|_2$ denote the normalized representation. The diversity loss is defined as
\begin{equation}
\mathcal L_{\mathrm{div}}
=
\frac{2}{V(V-1)}
\sum_{1\leq i<j\leq V}
\left[
\left(
\widehat{\mathbf z}^{\,i}
\right)^{\top}
\widehat{\mathbf z}^{\,j}
\right]^2.
\end{equation}

Minimizing this loss reduces the similarity between private-expert representations and encourages the experts to preserve distinct view-specific information.

\noindent\textbf{Correlation-aware weight regularization.}
Although the preceding evidence correction mechanism suppresses repeated 
information among private experts, similar expert evidence may still receive 
large weights simultaneously during the final fusion. We therefore further 
regularize the weight co-activation of all experts. The normalized evidence 
similarity between experts $i$ and $j$ is defined as
\begin{equation}
r_{ij}
=
1-
\frac{
\left\|
\mathbf e^{(i)}-\mathbf e^{(j)}
\right\|_1
}{
\left\|\mathbf e^{(i)}\right\|_1
+
\left\|\mathbf e^{(j)}\right\|_1
}.
\end{equation}

Based on this similarity, the correlation-aware weight regularizer is 
formulated as
\begin{equation}
\mathcal L_{\mathrm{red}}
=
\sum_{1\leq i<j\leq V+1}
r_{ij}w_iw_j.
\end{equation}

Here, $w_iw_j$ measures the weight co-activation of experts $i$ and $j$. 
When two experts produce similar evidence and simultaneously receive large 
weights, this regularizer imposes a stronger penalty, thereby preventing 
correlated evidence from being repeatedly emphasized in the final fusion.

The overall training objective is
\begin{equation}
\mathcal L
=
\mathcal L_{\mathrm{fuse}}
+
\lambda\mathcal L_{\mathrm{exp}}
+
\gamma\mathcal L_{\mathrm{div}}
+
\eta\mathcal L_{\mathrm{red}},
\end{equation}
where $\lambda$, $\gamma$, and $\eta$ control the contributions of average 
expert supervision, private-expert diversity regularization, and 
correlation-aware weight regularization, respectively.

\begin{table*}[t]
\centering
\begin{tabular}{lcccccc}
\toprule
Method & HandWritten & NUS-WIDE & Scene15 & MSRCV1 & 100Leaves & YouTube \\
\midrule
UVF
& $94.7\pm8.0$ & $\underline{73.4\pm0.6}$ & $\underline{74.7\pm1.6}$
& $96.2\pm3.2$ & $\underline{99.7\pm0.3}$ & $\underline{88.1\pm1.4}$ \\
AssoDMVC
& $94.0\pm10.0$ & $71.8\pm0.9$ & $54.4\pm27.8$
& $91.4\pm3.9$ & $89.1\pm13.3$ & $78.4\pm11.8$ \\
TMC
& $97.2\pm1.1$ & $72.1\pm0.4$ & $66.4\pm1.0$
& $95.2\pm2.6$ & $87.6\pm1.6$ & $81.0\pm1.3$ \\
ECML
& $97.0\pm1.1$ & $72.2\pm0.4$ & $66.1\pm1.2$
& $95.7\pm2.3$ & $91.8\pm1.2$ & $81.3\pm1.4$ \\
CCML
& $96.6\pm1.0$ & $71.8\pm0.4$ & $66.2\pm1.1$
& $92.4\pm3.2$ & $86.0\pm1.8$ & $81.7\pm1.1$ \\
RCMCL
& $97.1\pm1.0$ & $69.9\pm0.5$ & $65.8\pm1.0$
& $95.7\pm1.8$ & $73.2\pm1.9$ & $83.0\pm1.3$ \\
SAEML
& $96.8\pm0.7$ & $55.6\pm1.2$ & $58.9\pm1.1$
& $95.7\pm2.3$ & $93.3\pm0.3$ & $81.2\pm1.8$ \\
TMCEK
& $97.2\pm1.0$ & $68.8\pm0.5$ & $67.7\pm1.0$
& $83.8\pm4.1$ & $91.2\pm1.6$ & $85.0\pm1.5$ \\
TUNED
& $\underline{98.0\pm0.8}$ & $57.0\pm1.2$ & $58.3\pm4.3$
& $\underline{96.7\pm1.2}$ & $84.8\pm2.0$ & $74.6\pm1.3$ \\
ETF
& $97.3\pm0.7$ & $\underline{73.4\pm0.4}$ & $69.6\pm1.7$
& $95.7\pm1.8$ & $98.3\pm0.6$ & $82.0\pm1.4$ \\
HOVQ
& $96.9\pm1.1$ & $71.5\pm0.5$ & $66.6\pm1.1$
& $95.2\pm2.6$ & $84.4\pm1.7$ & $78.8\pm1.1$ \\
\midrule
CAFE (Ours)
& $\mathbf{98.6\pm0.5}$ & $\mathbf{73.6\pm0.5}$ & $\mathbf{82.1\pm1.4}$
& $\mathbf{98.6\pm1.9}$ & $\mathbf{99.8\pm0.2}$ & $\mathbf{88.3\pm2.0}$ \\
\bottomrule
\end{tabular}
\caption{Classification accuracy (\%) on six multi-view datasets. The best and second-best results are highlighted in bold and underlined, respectively.}
\label{tab:accuracy_comparison}
\end{table*}

\section{Experiments}

\subsection{Experimental Setup}

\subsubsection{Datasets}

We conduct experiments on six publicly available multi-view benchmark datasets, including \textbf{HandWritten}, \textbf{NUS-WIDE-128}, \textbf{Scene15}, \textbf{MSRCV1}, \textbf{100Leaves}, and \textbf{YouTube}. Detailed descriptions of these datasets are provided in the Appendix.

\subsubsection{Compared Methods}

To comprehensively evaluate the proposed method, we compare it with representative multi-view classification methods from two categories. Feature-level fusion methods include \textbf{UVF}~\cite{lv2026uvf} and \textbf{AssoDMVC}~\cite{liang2025view}. Trusted multi-view classification methods include \textbf{TMC}~\cite{han2021trusted}, \textbf{ECML}~\cite{xu2024reliable}, \textbf{CCML}~\cite{liu2024dynamic}, \textbf{RCMCL}~\cite{hu2026robust}, \textbf{SAEML}~\cite{xu2025beyond}, \textbf{TMCEK}~\cite{liang2025trusted}, \textbf{TUNED}~\cite{huang2025tuned}, \textbf{ETF}~\cite{lu2025navigating}, and \textbf{HOVQ}~\cite{hu2026hovq}. Details of the compared methods are provided in the Appendix.

\subsubsection{Implementation Details}

All datasets are split into training and test sets using stratified random sampling with a ratio of 8:2. The model is implemented in PyTorch, with the unified feature dimension, expert hidden dimension, and gating network hidden dimension set to 256, 512, and 128, respectively. We use the Adam optimizer with an initial learning rate of $10^{-3}$. All experiments are independently repeated with five random seeds, and the mean and standard deviation of accuracy and ECE are reported.

\begin{table*}[t]
\centering
\begin{tabular}{@{}lcccccc@{}}
\toprule
Method & HandWritten & NUS-WIDE & Scene15 & MSRCV1 & 100Leaves & YouTube \\
\midrule
TMC
& $15.9\pm1.2$ & $31.8\pm1.2$ & $46.2\pm1.0$
& $\underline{20.4\pm2.3}$ & $81.5\pm1.7$ & $26.1\pm1.5$ \\
ECML
& $60.2\pm1.2$ & $54.9\pm1.0$ & $54.3\pm1.2$
& $64.4\pm2.2$ & $88.4\pm1.2$ & $60.6\pm1.6$ \\
CCML
& $47.6\pm1.1$ & $43.2\pm0.2$ & $46.9\pm1.1$
& $41.9\pm2.6$ & $81.5\pm1.8$ & $48.4\pm0.9$ \\
RCMCL
& $34.8\pm1.0$ & $35.2\pm0.6$ & $47.2\pm1.0$
& $31.3\pm1.6$ & $68.8\pm1.9$ & $33.9\pm1.3$ \\
SAEML
& $\underline{0.9\pm0.4}$ & $42.0\pm1.2$ & $\underline{34.9\pm1.1}$
& $\mathbf{2.1\pm2.3}$ & $\underline{18.2\pm0.4}$ & $\underline{15.5\pm1.8}$ \\
TMCEK
& $68.7\pm1.0$ & $34.9\pm0.9$ & $53.0\pm1.1$
& $39.9\pm4.8$ & $86.0\pm1.6$ & $56.0\pm1.5$ \\
TUNED
& $50.7\pm1.1$ & $38.5\pm6.3$ & $40.3\pm4.7$
& $54.4\pm3.7$ & $74.4\pm4.2$ & $44.8\pm1.4$ \\
ETF
& $15.5\pm1.5$ & $\underline{25.7\pm0.8}$ & $51.4\pm1.7$
& $25.7\pm1.7$ & $71.9\pm0.6$ & $28.1\pm1.2$ \\
HOVQ
& $25.9\pm1.1$ & $38.9\pm1.1$ & $49.7\pm1.1$
& $33.4\pm1.9$ & $79.1\pm1.7$ & $33.1\pm1.3$ \\
\midrule
CAFE (Ours)
& $\mathbf{0.6\pm0.5}$ & $\mathbf{9.3\pm2.8}$ & $\mathbf{7.8\pm1.6}$
& $\mathbf{2.1\pm0.9}$ & $\mathbf{12.7\pm0.2}$ & $\mathbf{9.5\pm1.8}$ \\
\bottomrule
\end{tabular}
\caption{Calibration results on trusted multi-view classification methods (ECE\%, lower is better).}
\label{tab:ece_comparison}
\end{table*}

\begin{table}[t]
\centering
\small
\begin{tabular}{lccc}
\toprule
Variant & HandWritten & YouTube & 100Leaves \\
\midrule
w/o Evid.\ Corr.
& $\mathbf{98.6\pm0.6}$ & $86.4\pm1.0$ & $90.6\pm3.8$ \\

w/o $\mathcal{L}_{\mathrm{red}}$
& $\mathbf{98.6\pm0.8}$ & $81.4\pm13.3$ & $99.4\pm0.3$ \\

w/o $\mathcal{L}_{\mathrm{div}}$
& $\mathbf{98.6\pm0.5}$ & $88.0\pm1.5$ & $99.6\pm0.3$ \\
\midrule

Full method
& $\mathbf{98.6\pm0.5}$ & $\mathbf{88.3\pm2.0}$
& $\mathbf{99.8\pm0.2}$ \\
\bottomrule
\end{tabular}
\caption{Ablation study of different components in CAFE in terms of classification accuracy (\%).}
\label{tab:ablation}
\end{table}

\subsection{Comparison Results}

Table~\ref{tab:accuracy_comparison} reports the classification accuracy on six datasets, with results given as the mean and standard deviation. CAFE achieves the highest accuracy on all datasets. In particular, it outperforms the second-best methods by 7.4\% and 1.9\% on Scene15 and MSRCV1, respectively. Across the six datasets, CAFE achieves an average accuracy of 90.2\%, exceeding UVF, the strongest baseline by average accuracy, by 2.4\%. These results demonstrate that CAFE effectively exploits complementary information across views while maintaining stable classification performance across different types of datasets.

Classification accuracy alone is insufficient to evaluate a trusted multi-view classification model, as predicted confidence should be well aligned with actual predictive performance. Therefore, Table~\ref{tab:ece_comparison} compares the ECE values of trusted multi-view classification methods. CAFE achieves the lowest ECE on five datasets and ties for best on MSRCV1. In particular, CAFE reduces ECE by 16.4\% and 27.1\% on NUS-WIDE and Scene15, respectively, compared with the best-performing baselines. On average, CAFE achieves an ECE of 7.0\%, reducing the 18.9\% average ECE of SAEML, the best baseline, by 11.9\%. These results indicate that the improvements of CAFE are not limited to classification accuracy; its predictive confidence is also substantially better calibrated and more reliable.

\begin{figure}[t]
\centering
\includegraphics[width=\columnwidth]{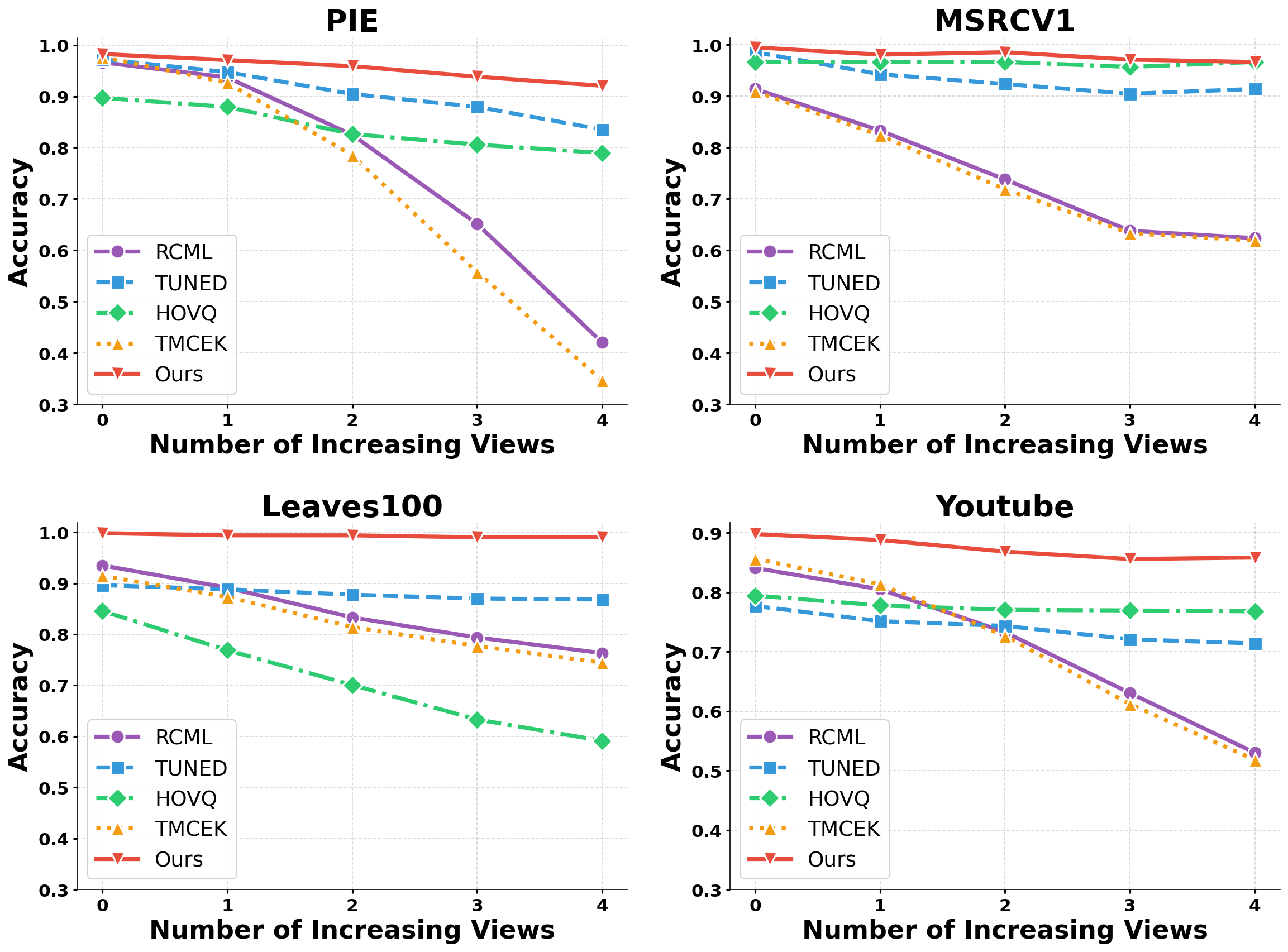}
\caption{Classification accuracy under strong view correlation constructed by progressively adding replicated views.}
\label{fig:view_replication}
\end{figure}

\subsection{Further Analysis}

\subsubsection{Impact of View Correlation}

To investigate the impact of strong view correlation, we construct correlated variants of four real-world datasets by progressively adding replicated views from the original view set. Since the replicated views provide no additional information, an ideal fusion method should remain stable as their number increases. Figure~\ref{fig:view_replication} shows the classification accuracy as the number of replicated views increases from zero to four.

Most comparison methods suffer noticeable performance degradation as more replicated views are added. With four additional replicated views, the average accuracy of RCML and TMCEK decreases by 33.0\% and 35.7\%, respectively, while TUNED and HOVQ also exhibit decreases of 7.5\% and 9.7\%. In contrast, the average accuracy of CAFE decreases by only 3.4\%. Specifically, its accuracy drops by 6.2\%, 2.9\%, 0.8\%, and 4.0\% on PIE, MSRCV1, 100Leaves, and YouTube, respectively, remaining nearly unchanged on 100Leaves. Under the strongest correlation setting, CAFE achieves the best or tied-best accuracy on all four datasets. These results demonstrate that CAFE effectively suppresses the repeated influence of strongly correlated views and maintains greater stability as view correlation increases.

\begin{figure}[t]
\centering
\includegraphics[width=\columnwidth]{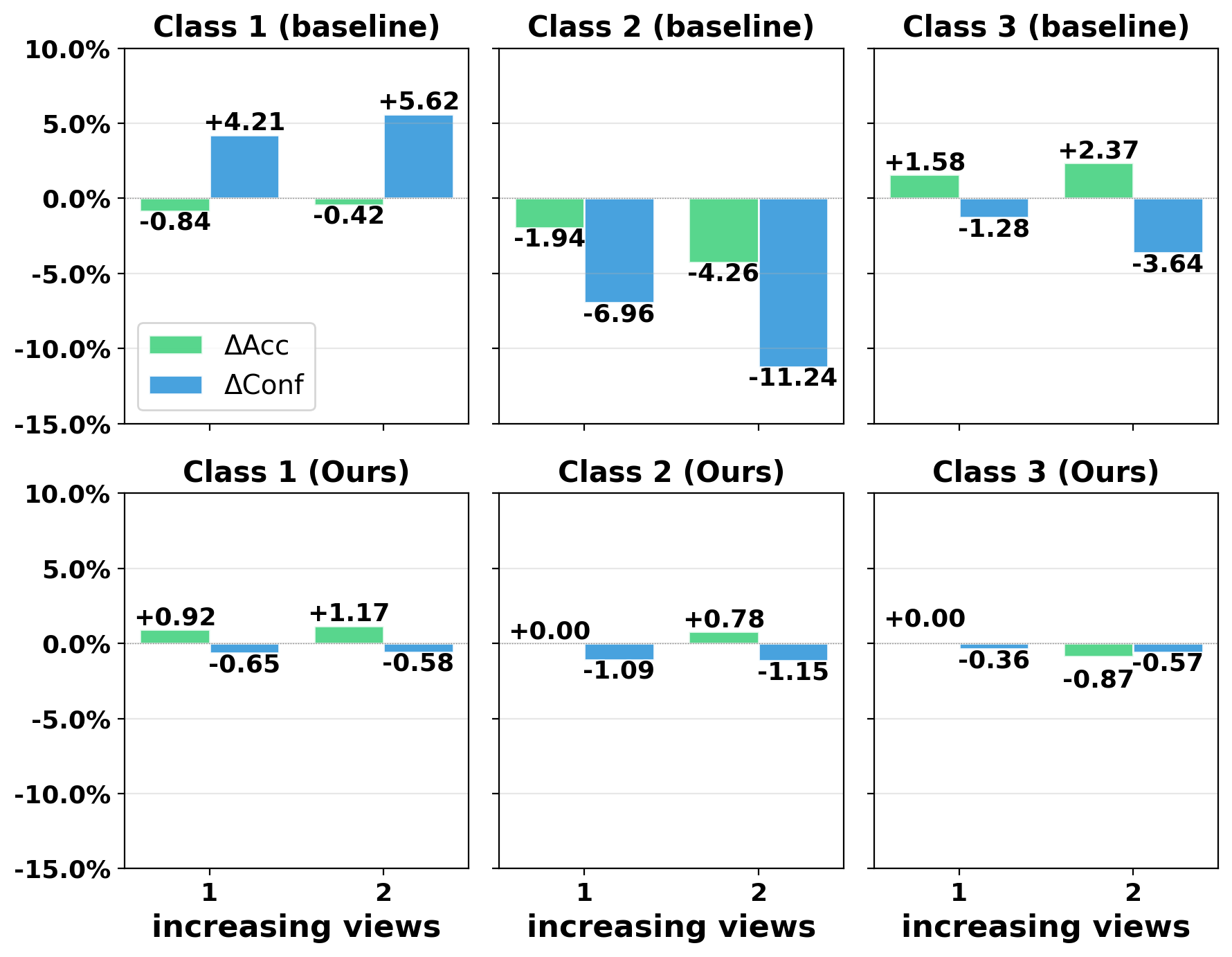}
\caption{Class-wise changes in accuracy ($\Delta$Acc) and confidence ($\Delta$Conf) as highly correlated views are added.}
\label{fig:toy_uncertainty}
\end{figure}

\subsubsection{Uncertainty under Strong View Correlation}

To examine the effect of highly correlated views on uncertainty estimation, we construct a three-class toy dataset, detailed in the Appendix. View~$A$ distinguishes Class~1, View~$C$ provides complementary information for Class~2, and View~$B$ is highly correlated with View~$A$. Starting from $A+C$, we add one or two copies of View~$B$ to form $A+C+B$ and $A+C+B+B$. The baseline averages the view outputs, and we report class-wise changes in accuracy and confidence relative to $A+C$.

As shown in Figure~\ref{fig:toy_uncertainty}, adding one and two correlated views changes the baseline accuracy on Class~1 by $-0.84\%$ and $-0.42\%$, but increases its confidence by $4.21\%$ and $5.62\%$, respectively. This indicates that correlated views inflate confidence without providing additional discriminative information, shifting predictions toward overconfidence. For Class~2, accuracy decreases by $1.94\%$ and $4.26\%$, whereas confidence drops by $6.96\%$ and $11.24\%$, suggesting that correlated views weaken complementary information and shift predictions toward underconfidence. In contrast, CAFE keeps both accuracy and confidence changes within approximately $1.2\%$ across all classes, demonstrating stable uncertainty estimation by suppressing repeated correlated evidence.

\subsubsection{Ablation Study}

As shown in Table~\ref{tab:ablation}, the complete CAFE model achieves the highest average accuracy of 95.6\% across the three datasets. The variants without evidence correction, $\mathcal{L}_{\mathrm{red}}$, and $\mathcal{L}_{\mathrm{div}}$ obtain average accuracies of 91.9\%, 93.1\%, and 95.4\%, respectively. Removing evidence correction decreases accuracy on YouTube and 100Leaves by 1.9\% and 9.2\%, while removing $\mathcal{L}_{\mathrm{red}}$ causes a 6.9\% decrease on YouTube and increases the standard deviation from 2.0\% to 13.3\%. The complete model achieves the best overall performance, demonstrating the effectiveness of the three components.


\section{Conclusion}

In this paper, we proposed CAFE, a correlation-aware evidential fusion framework for trusted multi-view classification. CAFE constructs view-private and cross-view collaborative experts to capture view-specific and joint decision information. It further corrects private evidence according to inter-view complementarity, suppressing repeated contributions from correlated views while preserving complementary information. A global context-based weighting mechanism then adaptively integrates the corrected private and collaborative evidence for each sample. Experiments on six real-world datasets confirmed the effectiveness of CAFE in classification and uncertainty estimation. Further experiments in highly correlated settings demonstrated its robustness and calibration stability.




\bibliography{aaai2027}

@inproceedings{wei2026unbiased,
  title={Unbiased dynamic multimodal fusion},
  author={Wei, Shicai and Zhang, Kaijie and Chen, Luyi and He, Tao and Duan, Guiduo},
  booktitle={Proceedings of the IEEE/CVF Conference on Computer Vision and Pattern Recognition},
  pages={6239--6249},
  year={2026}
}

@inproceedings{he2026bootstrapping,
  title={Bootstrapping multi-view learning for test-time noisy correspondence},
  author={He, Changhao and Xue, Di and Li, Shuxian and Hao, Yanji and Peng, Xi and Hu, Peng},
  booktitle={Proceedings of the IEEE/CVF Conference on Computer Vision and Pattern Recognition},
  pages={1627--1638},
  year={2026}
}

@inproceedings{wei2023surroundocc,
  title={Surroundocc: Multi-camera 3d occupancy prediction for autonomous driving},
  author={Wei, Yi and Zhao, Linqing and Zheng, Wenzhao and Zhu, Zheng and Zhou, Jie and Lu, Jiwen},
  booktitle={Proceedings of the IEEE/CVF International Conference on Computer Vision},
  pages={21729--21740},
  year={2023}
}

@article{chen2022pathomic,
  author  = {Chen, Richard J. and Lu, Ming Y. and Wang, Jingwen and Williamson, Drew F. K. and Rodig, Scott J. and Lindeman, Neal I. and Mahmood, Faisal},
  title   = {Pathomic Fusion: An Integrated Framework for Fusing Histopathology and Genomic Features for Cancer Diagnosis and Prognosis},
  journal = {IEEE Transactions on Medical Imaging},
  volume  = {41},
  number  = {4},
  pages   = {757--770},
  year    = {2022},
  doi     = {10.1109/TMI.2020.3021387}
}

@article{chen2022pancancer,
  author  = {Chen, Richard J. and Lu, Ming Y. and Williamson, Drew F. K. and Chen, Tiffany Y. and Lipkova, Jana and Noor, Zahra and Shaban, Muhammad and Shady, Maha and Williams, Mane and Joo, Bumjin and Mahmood, Faisal},
  title   = {Pan-Cancer Integrative Histology-Genomic Analysis via Multimodal Deep Learning},
  journal = {Cancer Cell},
  volume  = {40},
  number  = {8},
  pages   = {865--878.e6},
  year    = {2022},
  doi     = {10.1016/j.ccell.2022.07.004}
}

@inproceedings{clark2026molmo2,
  title={Molmo2: Open weights and data for vision-language models with video understanding and grounding},
  author={Clark, Christopher and Zhang, Jieyu and Ma, Zixian and Park, Jae Sung and Tripathi, Rohun and Lee, Sangho and Salehi, Mohammadreza and Ren, Jason and Kim, Chris Dongjoo and Yang, Yinuo and others},
  booktitle={Proceedings of the IEEE/CVF Conference on Computer Vision and Pattern Recognition},
  pages={28652--28668},
  year={2026}
}

@inproceedings{xu2024labelnoise,
  author    = {Xu, Cai and Zhang, Yilin and Guan, Ziyu and Zhao, Wei},
  title     = {Trusted {Multi-View} Learning with Label Noise},
  booktitle = {Proceedings of the Thirty-Third International Joint Conference on Artificial Intelligence},
  pages     = {5263--5271},
  year      = {2024},
  doi       = {10.24963/ijcai.2024/582}
}

@inproceedings{xie2023exploring,
  author    = {Xie, Mengyao and Han, Zongbo and Zhang, Changqing and Bai, Yichen and Hu, Qinghua},
  title     = {Exploring and Exploiting Uncertainty for Incomplete {Multi-View} Classification},
  booktitle = {Proceedings of the IEEE/CVF Conference on Computer Vision and Pattern Recognition},
  pages     = {19873--19882},
  year      = {2023}
}

@inproceedings{liu2025enhancing,
  title     = {Enhancing Testing-Time Robustness for Trusted Multi-View
               Classification in the Wild},
  author    = {Liu, Wei and Chen, Yufei and Yue, Xiaodong},
  booktitle = {Proceedings of the IEEE/CVF Conference on Computer Vision
               and Pattern Recognition},
  pages     = {15508--15517},
  year      = {2025}
}

@incollection{hotelling1992relations,
  title={Relations between two sets of variates},
  author={Hotelling, Harold},
  booktitle={Breakthroughs in statistics: methodology and distribution},
  pages={162--190},
  year={1992},
  publisher={Springer}
}

@inproceedings{andrew2013deep,
  author    = {Andrew, Galen and Arora, Raman and Bilmes, Jeff and Livescu, Karen},
  title     = {Deep Canonical Correlation Analysis},
  booktitle = {Proceedings of the 30th International Conference on Machine Learning},
  series    = {Proceedings of Machine Learning Research},
  volume    = {28},
  number    = {3},
  pages     = {1247--1255},
  year      = {2013},
  publisher = {PMLR}
}

@inproceedings{wang2015deep,
  author    = {Wang, Weiran and Arora, Raman and Livescu, Karen and Bilmes, Jeff},
  title     = {On Deep {Multi-View} Representation Learning},
  booktitle = {Proceedings of the 32nd International Conference on Machine Learning},
  series    = {Proceedings of Machine Learning Research},
  volume    = {37},
  pages     = {1083--1092},
  year      = {2015},
  publisher = {PMLR}
}

@inproceedings{benton2019deep,
  author    = {Benton, Adrian and Khayrallah, Huda and Gujral, Biman and Reisinger, Dee Ann and Zhang, Sheng and Arora, Raman},
  title     = {Deep Generalized Canonical Correlation Analysis},
  booktitle = {Proceedings of the 4th Workshop on Representation Learning for NLP},
  pages     = {1--6},
  year      = {2019},
  publisher = {Association for Computational Linguistics},
  doi       = {10.18653/v1/W19-4301}
}

@article{xu2015multiview,
  author  = {Xu, Chang and Tao, Dacheng and Xu, Chao},
  title   = {{Multi-View} Intact Space Learning},
  journal = {IEEE Transactions on Pattern Analysis and Machine Intelligence},
  volume  = {37},
  number  = {12},
  pages   = {2531--2544},
  year    = {2015},
  doi     = {10.1109/TPAMI.2015.2417578}
}

@inproceedings{xu2021multivae,
  author    = {Xu, Jie and Ren, Yazhou and Tang, Huayi and Pu, Xiaorong and Zhu, Xiaofeng and Zeng, Ming and He, Lifang},
  title     = {{Multi-VAE}: Learning Disentangled View-Common and View-Peculiar Visual Representations for {Multi-View} Clustering},
  booktitle = {Proceedings of the IEEE/CVF International Conference on Computer Vision},
  pages     = {9234--9243},
  year      = {2021}
}

@inproceedings{liu2024learn,
  author    = {Liu, Suyuan and Liang, Ke and Dong, Zhibin and Wang, Siwei and Yang, Xihong and Zhou, Sihang and Zhu, En and Liu, Xinwang},
  title     = {Learn from View Correlation: An Anchor Enhancement Strategy for Multi-View Clustering},
  booktitle = {Proceedings of the IEEE/CVF Conference on Computer Vision and Pattern Recognition},
  pages     = {26151--26161},
  year      = {2024}
}

@article{chen2025dfl,
  author  = {Chen, Zhe and Wu, Xiao-Jun and Xu, Tianyang and Kittler, Josef},
  title   = {{DFL-Net}: Disentangled Feature Learning Network for Multi-View Clustering},
  journal = {IEEE Transactions on Knowledge and Data Engineering},
  volume  = {37},
  number  = {8},
  pages   = {4537--4547},
  year    = {2025},
  doi     = {10.1109/TKDE.2025.3574150}
}

@article{li2026miscc,
  author  = {Li, Pengyuan and Chang, Dongxia and Wang, Yiming and Liu, Man and Kong, Zisen and Kong, Linhua and Zhao, Yao},
  title   = {Deep Multi-View Clustering With Intra-View Similarity and Cross-View Correlation Learning},
  journal = {IEEE Transactions on Multimedia},
  volume  = {28},
  pages   = {3682--3693},
  year    = {2026},
  doi     = {10.1109/TMM.2026.3654390}
}

@inproceedings{dong2023crossview,
  author    = {Dong, Zhibin and Wang, Siwei and Jin, Jiaqi and Liu, Xinwang and Zhu, En},
  title     = {Cross-View Topology Based Consistent and Complementary Information for Deep {Multi-View} Clustering},
  booktitle = {Proceedings of the IEEE/CVF International Conference on Computer Vision},
  pages     = {19440--19451},
  year      = {2023}
}

@inproceedings{liang2025view,
  author    = {Liang, Xinyan and Lv, Li and Guo, Qian and Jiang, Bingbing and Li, Feijiang and Du, Liang and Chen, Lu},
  title     = {View-Association-Guided Dynamic {Multi-View} Classification},
  booktitle = {Proceedings of the Thirty-Fourth International Joint Conference on Artificial Intelligence},
  pages     = {5680--5688},
  year      = {2025},
  doi       = {10.24963/ijcai.2025/632}
}

@article{gawlikowski2023survey,
  author  = {Gawlikowski, Jakob and Tassi, Cedrique Rovile Njieutcheu and Ali, Mohsin and Lee, Jongseok and Humt, Matthias and Feng, Jianxiang and Kruspe, Anna and Triebel, Rudolph and Jung, Peter and Roscher, Ribana and Shahzad, Muhammad and Yang, Wen and Bamler, Richard and Zhu, Xiao Xiang},
  title   = {A Survey of Uncertainty in Deep Neural Networks},
  journal = {Artificial Intelligence Review},
  volume  = {56},
  number  = {Suppl 1},
  pages   = {S1513--S1589},
  year    = {2023},
  doi     = {10.1007/s10462-023-10562-9}
}

@inproceedings{sensoy2018evidential,
  author    = {Sensoy, Murat and Kaplan, Lance and Kandemir, Melih},
  title     = {Evidential Deep Learning to Quantify Classification Uncertainty},
  booktitle = {Advances in Neural Information Processing Systems},
  volume    = {31},
  year      = {2018}
}

@inproceedings{han2021trusted,
  author    = {Han, Zongbo and Zhang, Changqing and Fu, Huazhu and Zhou, Joey Tianyi},
  title     = {Trusted {Multi-View} Classification},
  booktitle = {International Conference on Learning Representations},
  year      = {2021}
}

@article{xu2024reliable,
  author  = {Xu, Cai and Si, Jiajun and Guan, Ziyu and Zhao, Wei and Wu, Yue and Gao, Xiyue},
  title   = {Reliable Conflictive {Multi-View} Learning},
  journal = {Proceedings of the AAAI Conference on Artificial Intelligence},
  volume  = {38},
  number  = {14},
  pages   = {16129--16137},
  year    = {2024},
  doi     = {10.1609/aaai.v38i14.29546}
}

@inproceedings{xu2025beyond,
  author    = {Xu, Cai and Wen, Ziqi and Zhao, Jie and Zhao, Wanqing and Yu, Jinlong and Chen, Haishun and Guan, Ziyu and Zhao, Wei},
  title     = {Beyond Equal Views: Strength-Adaptive Evidential {Multi-View} Learning},
  booktitle = {Proceedings of the 33rd ACM International Conference on Multimedia},
  pages     = {1278--1287},
  year      = {2025},
  publisher = {Association for Computing Machinery},
  doi       = {10.1145/3746027.3755092}
}

@inproceedings{liang2025trusted,
  author    = {Liang, Xinyan and Wang, Shijie and Qian, Yuhua and Guo, Qian and Du, Liang and Jiang, Bingbing and Luo, Tingjin and Li, Feijiang},
  title     = {Trusted {Multi-View} Classification with Expert Knowledge Constraints},
  booktitle = {Proceedings of the 42nd International Conference on Machine Learning},
  series    = {Proceedings of Machine Learning Research},
  volume    = {267},
  pages     = {37409--37426},
  year      = {2025},
  publisher = {PMLR}
}

@article{zhang2026independently,
  author        = {Zhang, Yilin and Xu, Cai and Chen, Haishun and Guan, Ziyu and Zhao, Wei},
  title         = {Are Independently Estimated View Uncertainties Comparable? Unified Routing for Trusted {Multi-View} Classification},
  journal       = {arXiv preprint arXiv:2604.09288},
  year          = {2026},
  eprint        = {2604.09288},
  archivePrefix = {arXiv},
  primaryClass  = {cs.LG},
  doi           = {10.48550/arXiv.2604.09288}
}

@inproceedings{liu2024dynamic,
  title     = {Dynamic Evidence Decoupling for Trusted Multi-View Learning},
  author    = {Liu, Ying and Liu, Lihong and Xu, Cai and Song, Xiangyu and Guan, Ziyu and Zhao, Wei},
  booktitle = {Proceedings of the 32nd ACM International Conference on Multimedia},
  pages     = {7269--7277},
  year      = {2024},
  doi       = {10.1145/3664647.3681404}
}

@inproceedings{hu2026hovq,
  title     = {Hyper-Opinion Vagueness Quantification for Robust Multimodal Learning},
  author    = {Hu, Disen and Jiang, Xun and Cao, Xiaofeng and Wang, Zheng and Song, Jingkuan and Shen, Heng Tao and Xu, Xing},
  booktitle = {Proceedings of the AAAI Conference on Artificial Intelligence},
  volume    = {40},
  number    = {26},
  pages     = {21831--21839},
  year      = {2026},
  doi       = {10.1609/aaai.v40i26.39335}
}

@inproceedings{lv2026uvf,
  author    = {Lv, Li and Guo, Qian and Zhang, Li and Du, Liang and Jiang, Bingbing and Chen, Lu and Liang, Xinyan},
  title     = {Uncertainty-Guided View-Strength-Aware Feature Utilization for {Multi-View} Classification},
  booktitle = {Proceedings of the AAAI Conference on Artificial Intelligence},
  volume    = {40},
  number    = {29},
  pages     = {24198--24206},
  year      = {2026},
  doi       = {10.1609/aaai.v40i29.39599}
}

@article{hu2026robust,
  author  = {Hu, Shaobo and Huang, Hui and Zhang, Nan and Sun, Shiliang},
  title   = {Robust Trusted Conflictive Multiview Collaborative Contrastive Learning},
  journal = {IEEE Transactions on Pattern Analysis and Machine Intelligence},
  volume  = {48},
  number  = {6},
  pages   = {7161--7176},
  year    = {2026},
  doi     = {10.1109/TPAMI.2026.3663788}
}

@inproceedings{huang2025tuned,
  author    = {Huang, Haojian and Qin, Chuanyu and Liu, Zhe and Ma, Kaijing and Chen, Jin and Fang, Han and Ban, Chao and Sun, Hao and He, Zhongjiang},
  title     = {Trusted Unified Feature-Neighborhood Dynamics for {Multi-View} Classification},
  booktitle = {Proceedings of the AAAI Conference on Artificial Intelligence},
  volume    = {39},
  number    = {16},
  pages     = {17413--17421},
  year      = {2025},
  doi       = {10.1609/aaai.v39i16.33914}
}

@inproceedings{lu2025navigating,
  author    = {Lu, Jueqing and Buntine, Wray and Qi, Yuanyuan and Dipnall, Joanna and Gabbe, Belinda and Du, Lan},
  title     = {Navigating Conflicting Views: Harnessing Trust for Learning},
  booktitle = {Proceedings of the 42nd International Conference on Machine Learning},
  series    = {Proceedings of Machine Learning Research},
  volume    = {267},
  pages     = {40411--40435},
  publisher = {PMLR},
  year      = {2025}
}


\clearpage
\appendix
\section*{Appendix}

\setcounter{figure}{0}
\setcounter{table}{0}
\setcounter{equation}{0}

\renewcommand{\thefigure}{S\arabic{figure}}
\renewcommand{\thetable}{S\arabic{table}}
\renewcommand{\theequation}{S\arabic{equation}}

\section{Experimental Settings}

\subsection{Datasets Details}

\textbf{HandWritten}\footnote{\url{https://archive.ics.uci.edu/dataset/72/multiple+features}} is a handwritten digit recognition dataset collected from the UCI Machine Learning Repository. It contains 2,000 samples from 10 digit classes, where each sample is represented by six feature views, including Fourier coefficients, profile correlations, Karhunen-Love coefficients, pixel averages, Zernike moments, and morphological features.

\textbf{NUS-WIDE-128}\footnote{\url{https://v2.imag-njust.net/resource/NUS-wide/}} is a large-scale web image classification dataset containing 43,800 image samples from 128 semantic categories. Each image is represented by seven complementary visual feature views.

\textbf{Scene15}\footnote{\url{https://doi.org/10.6084/m9.figshare.7007177.v1}} is a scene recognition dataset consisting of 4,485 images from 15 scene categories. Following previous multi-view learning studies, we represent each image using three feature views, including GIST, PHOG, and LBP features.

\textbf{MSRCV1}\footnote{\url{https://mldta.com/dataset/msrc-v1/}} is an object recognition dataset containing 210 images from seven object categories. Each image is described by six feature views.

\textbf{100Leaves}\footnote{\url{https://archive.ics.uci.edu/dataset/241/one+hundred+plant+species+leaves+data+set}} is a plant classification dataset containing 1,600 leaf samples from 100 plant species. The samples are represented by three feature views, including shape, margin, and texture features.

\textbf{YouTube}\footnote{\url{https://archive.ics.uci.edu/dataset/269/youtube+multiview+video+games+dataset}} is a multi-view video classification dataset collected from YouTube. We use 2,000 samples from 10 video game categories, represented by six views, including three visual views and three audio views.

\begin{figure*}[t]
\centering
\includegraphics[width=\textwidth]
{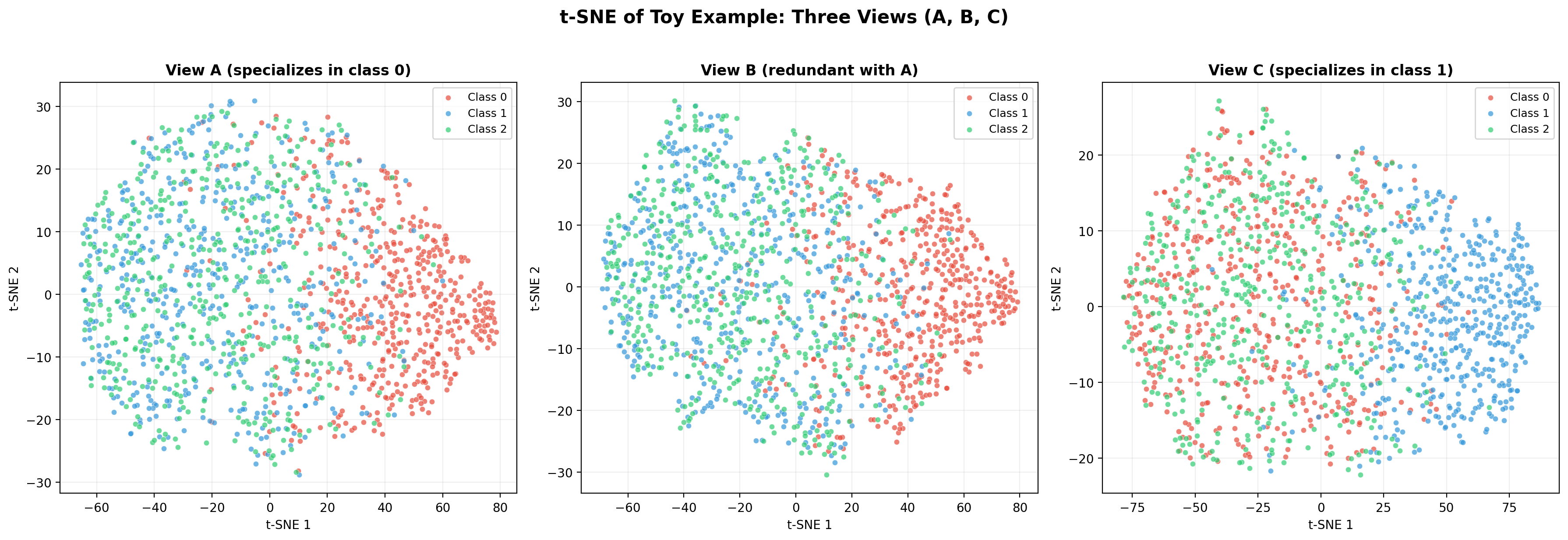}
\caption{Visualization of data instances in the toy dataset.}
\label{fig:toy_e}
\end{figure*}

\subsection{Compared Methods}

To comprehensively evaluate the proposed method, we select a diverse set of representative multi-view classification methods and divide them into two categories. The first category comprises feature-level multi-view fusion methods. \textbf{UVF}~\cite{lv2026uvf} evaluates the discriminative strength of each view through uncertainty estimation and accordingly performs feature selection and dynamic weighting. \textbf{AssoDMVC}~\cite{liang2025view} explicitly models inter-view associations and dynamically adjusts the contribution of each view to the fused representation. The second category comprises trusted multi-view classification methods. \textbf{TMC}~\cite{han2021trusted} models view-specific evidence using Dirichlet distributions and combines the resulting opinions through Dempster--Shafer theory. \textbf{ECML}~\cite{xu2024reliable} introduces an opinion aggregation strategy specifically designed for conflictive views. \textbf{CCML}~\cite{liu2024dynamic} dynamically decouples and integrates consistent and complementary evidence. \textbf{RCMCL}~\cite{hu2026robust} improves robustness under view conflicts through evidential contrastive learning and collaborative learning. \textbf{SAEML}~\cite{xu2025beyond} adaptively adjusts the fusion contribution of each view according to its evidence strength. \textbf{TMCEK}~\cite{liang2025trusted} incorporates expert knowledge constraints into trusted multi-view learning. \textbf{TUNED}~\cite{huang2025tuned} jointly models local and global feature-neighborhood structures. \textbf{ETF}~\cite{lu2025navigating} employs instance-wise trust discounting to reduce the influence of unreliable or conflicting opinions. \textbf{HOVQ}~\cite{hu2026hovq} uses hyper-opinions and composite classes to represent semantic vagueness and accordingly regulate the model optimization process.

\begin{table}[t]
\centering
\begin{tabular}{lcc}
\toprule
\textbf{Dataset} & $\boldsymbol{\gamma}$ & $\boldsymbol{\eta}$ \\
\midrule
HandWritten & 0     & 0.05  \\
NUS-WIDE   & 0.05  & 0.05  \\
Scene15    & 0.20  & 0.80  \\
MSRCV1     & 0.01  & 0.05  \\
100Leaves  & 0.002 & 10.00 \\
YouTube    & 0.01  & 0.05  \\
\bottomrule
\end{tabular}
\caption{Dataset-specific hyperparameters used by CAFE.}
\label{tab:supp_hyperparameters}
\end{table}

\begin{table}[t]
\centering
\resizebox{\columnwidth}{!}{
\begin{tabular}{lrcl}
\toprule
\textbf{Dataset} & \textbf{Size} & $\boldsymbol{K}$ &
\textbf{Dimensionality} \\
\midrule
HandWritten & 2,000  & 10  & 240/76/216/47/64/6 \\
NUS-WIDE   & 43,800 & 128 & 64/144/73/128/225/500/1000 \\
Scene15    & 4,485  & 15  & 20/59/40 \\
MSRCV1     & 210    & 7   & 1302/48/512/100/256/210 \\
100Leaves  & 1,600  & 100 & 64/64/64 \\
YouTube    & 2,000  & 10  & 2000/1024/64/512/64/647 \\
\bottomrule
\end{tabular}
}
\caption{Dataset summary.}
\label{tab:dataset_summary}
\end{table}

\subsection{Implementation Details}

All datasets are divided into training and test sets using an 80/20 stratified split. For each view, Min--Max normalization is fitted exclusively on the training set and then applied to the test set. All experiments are repeated using five random seeds, i.e., $\{3407, 7, 601, 101, 503\}$, and the mean and standard deviation are reported.

All experiments are conducted on a server running Ubuntu~18.04.6 LTS with Linux kernel~4.15.0-213, equipped with an Intel Core i9-10980XE CPU at 3.00\,GHz (18 cores and 36 threads), 125\,GB of system memory, and two NVIDIA TITAN RTX GPUs with 24\,GB of memory each. The CUDA and NVIDIA driver versions are 12.4 and 550.142, respectively. CAFE is implemented in Python~3.9.25 and PyTorch~2.6.0. The supporting libraries include NumPy~1.26.4, SciPy~1.13.1, and scikit-learn~1.6.1.

Each view is projected into a 256-dimensional feature space using a linear layer followed by layer normalization. Both the view-private experts and the cross-view collaborative expert contain a 512-dimensional hidden layer with ReLU activation and a dropout rate of 0.1. Softplus is used as the output activation to produce non-negative evidence. The hidden dimension of the global context-based weighting network is set to 128. During training, Gaussian noise with a standard deviation of 0.1 is added to the expert scores. The Softmax temperature is linearly annealed from 3.0 to 1.0 over the first 60\% of the training epochs. The uncertainty-weighted fusion operator is adopted to aggregate the discounted opinions during evidence correction.

CAFE is trained for 300 epochs with a batch size of 128 using the Adam optimizer. The initial learning rate and weight decay are set to $10^{-3}$ and $10^{-5}$, respectively. The learning rate is linearly warmed up during the first 10 epochs and subsequently decayed using a cosine schedule. Early stopping is not used. The weight of the average expert supervision $\lambda$ is fixed at 0.3, while the KL regularization term is disabled. The coefficients of the diversity and correlation-aware regularization terms are specified for each dataset, as shown in Table~\ref{tab:supp_hyperparameters}.

For uncertainty evaluation, ECE is computed using $1-u$ as the predictive confidence with 15 equally spaced bins.

\begin{figure*}[t]
\centering
\includegraphics[width=\textwidth]
{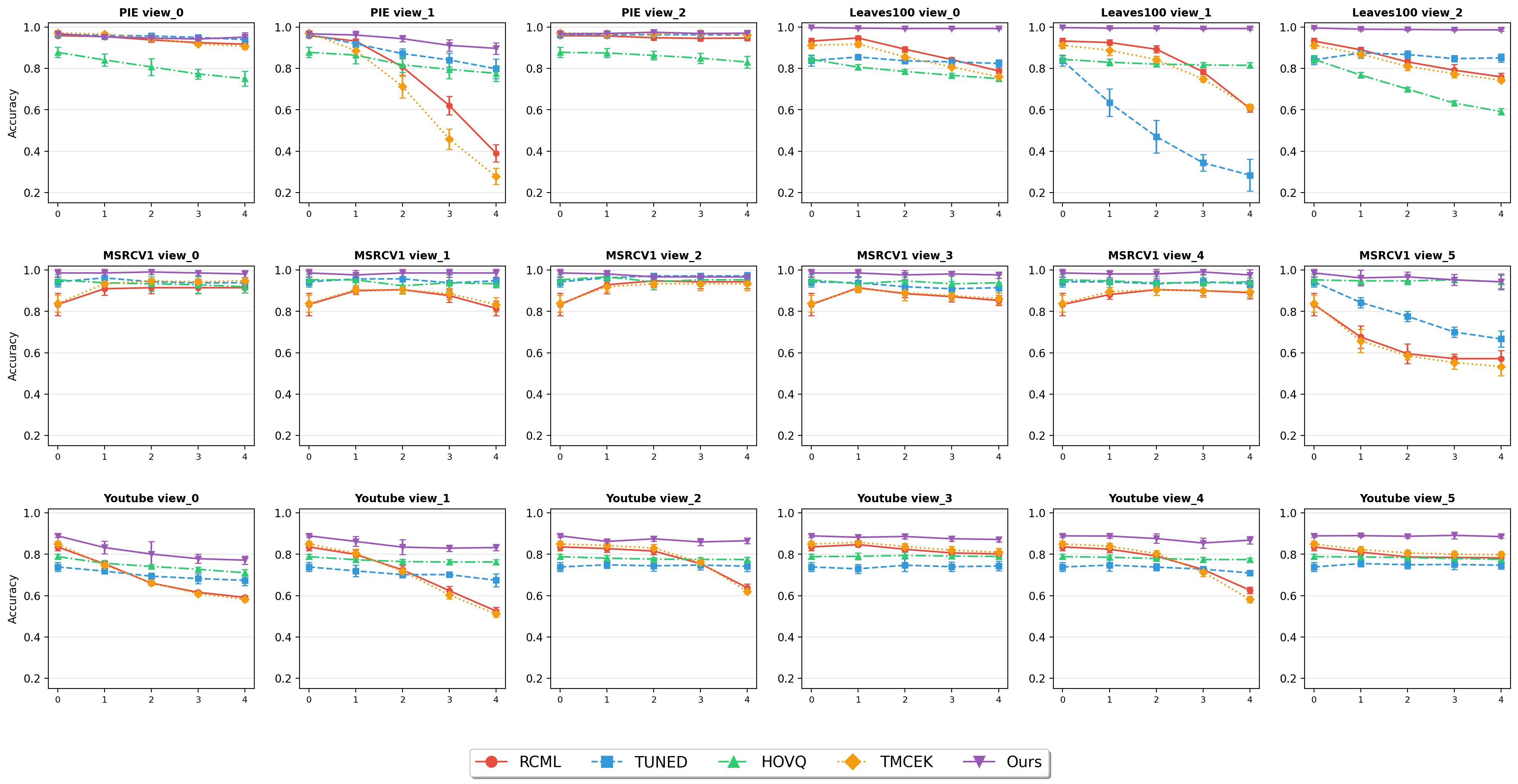}
\caption{View-replication results obtained by separately replicating
each original view. The first row presents the three views of PIE and
the three views of 100Leaves. The second and third rows present the six
views of MSRCV1 and YouTube, respectively.}
\label{fig:all_view_replication}
\end{figure*}

\subsection{Toy Dataset Construction}
\label{app:toy_dataset}

CAFE is designed to suppress repeated contributions from highly
correlated views while preserving complementary information. Therefore,
we construct a synthetic toy dataset to investigate how increasing view
correlation affects classification confidence and uncertainty
estimation.

The toy dataset contains 1,500 samples from three classes, with 500
samples per class. Each sample is first represented by a
six-dimensional latent vector, whose first and last three dimensions
encode two complementary factors. For Class~1, the first three
dimensions are sampled from $\mathcal{N}(0.6,0.4^{2})$, while the
remaining dimensions are sampled from $\mathcal{N}(0,0.4^{2})$. For
Class~2, the two distributions are reversed. For Class~3, all six
dimensions are sampled from $\mathcal{N}(0,0.4^{2})$. Consequently, the
first latent factor mainly distinguishes Class~1, the second factor
mainly distinguishes Class~2, and neither factor is specifically
informative for Class~3.

We generate three eight-dimensional views, denoted by $A$, $B$, and
$C$. Views~$A$ and~$C$ are obtained by projecting the first and second
latent factors, respectively. View~$B$ is generated from the same
latent factor as View~$A$ and is explicitly constructed to be highly
correlated with it:
\begin{equation}
\begin{aligned}
\mathbf{x}_{n}^{A}
&=
\mathbf{z}_{n}^{A}\mathbf{P}_{A}
+\boldsymbol{\epsilon}_{n}^{A},\\
\mathbf{x}_{n}^{B}
&=
\rho\mathbf{x}_{n}^{A}
+\sqrt{1-\rho^{2}}
\left(
\mathbf{z}_{n}^{A}\mathbf{P}_{B}
+\boldsymbol{\epsilon}_{n}^{B}
\right),\\
\mathbf{x}_{n}^{C}
&=
\mathbf{z}_{n}^{C}\mathbf{P}_{C}
+\boldsymbol{\epsilon}_{n}^{C},
\end{aligned}
\label{eq:toy_generation}
\end{equation}
where $\rho=0.99$ controls the correlation between Views~$A$ and~$B$.
The entries of
$\mathbf{P}_{A},\mathbf{P}_{B},\mathbf{P}_{C}\in\mathbb{R}^{3\times8}$
are sampled from $\mathcal{U}(0.4,1.0)$, with 20\% of them randomly set
to zero. The observation noise is sampled from
$\mathcal{N}(\mathbf{0},0.4^{2}\mathbf{I})$. Each generated view is
independently normalized to $[0,1]$.

Under this construction, Views~$A$ and~$B$ can distinguish Class~1 but
cannot effectively distinguish Classes~2 and~3. In contrast, View~$C$
can distinguish Class~2 but cannot effectively distinguish Classes~1
and~3. Therefore, Views~$A$ and~$B$ provide highly correlated
information, whereas View~$C$ provides complementary information.

We use $A+C$ as the initial setting and progressively add one and two
copies of View~$B$, resulting in $A+C+B$ and $A+C+B+B$, respectively.
The same View~$B$ is repeated in the last setting rather than generated
independently. Thus, the amount of correlated information increases
while the underlying samples and complementary information remain
unchanged. The samples are equally divided into training and evaluation
sets, and the changes in class-wise accuracy and confidence are
measured relative to the initial $A+C$ setting. All results are averaged
over five runs.

\section{Additional View-Replication Results}
\label{app:additional_view_replication}

In the main paper, we report the results obtained by replicating one
representative view from each dataset. To verify that the observations
are not specific to the selected view, we further repeat the experiment
for every original view of PIE, MSRCV1, 100Leaves, and YouTube. In each
experiment, the selected view is replicated one to four times, while
the remaining views and all experimental settings are kept unchanged.

As shown in Figure~\ref{fig:all_view_replication}, the degree of
performance degradation varies across replicated views, reflecting
differences in their discriminative information and relationships with
the remaining views. Nevertheless, the overall observation is
consistent: most comparison methods exhibit noticeable accuracy
degradation as correlated copies are added, whereas CAFE generally
maintains stable performance or shows only minor decreases. These
results confirm that the robustness of CAFE under strong view
correlation is not limited to a particular replicated view.

\end{document}